\documentclass[10pt,twocolumn,letterpaper]{article}

\usepackage{iccv}              

\definecolor{iccvblue}{rgb}{0.21,0.49,0.74}
\usepackage[pagebackref,breaklinks,colorlinks,allcolors=iccvblue]{hyperref}
\usepackage{amsmath}
\usepackage{letterspace}
\usepackage{setspace}

\newcommand{\Tref}[1]{Table~\ref{#1}}
\newcommand{\Fref}[1]{Fig.~\ref{#1}}

\def\paperID{337} 
\def\confName{ICCV}
\def\confYear{2025}

\title{Inverse Image-Based Rendering for \\ Light Field Generation from Single Images}

\author{Hyunjun Jung\\
GIST AI Graduated School\\
{\tt\small hyunjun.jung@gm.gist.ac.kr}
\and
Hae-Gon Jeon\\
Yonsei University\\
{\tt\small earboll@yonsei.ac.kr}
}

\begin{document}
\maketitle
\begin{abstract}
A concept of light-fields computed from multiple view images on regular grids has proven its benefit for scene representations, and supported realistic renderings of novel views and photographic effects such as refocusing and shallow depth of field. In spite of its effectiveness of light flow computations, obtaining light fields requires either computational costs or specialized devices like a bulky camera setup and a specialized microlens array.
  In an effort to broaden its benefit and applicability, in this paper, we propose a novel view synthesis method for light field generation from only single images, named \textit{inverse image-based rendering}. Unlike previous attempts to implicitly rebuild 3D geometry or to explicitly represent objective scenes, our method reconstructs light flows in a space from image pixels, which behaves in the opposite way to image-based rendering.
  To accomplish this, we design a neural rendering pipeline to render a target ray in an arbitrary viewpoint. Our neural renderer first stores the light flow of source rays from the input image, then computes the relationships among them through cross-attention, and finally predicts the color of the target ray based on these relationships.
  After the rendering pipeline generates the first novel view from a single input image, the generated out-of-view contents are updated to the set of source rays. This procedure is iteratively performed while ensuring the consistent generation of occluded contents.
  We demonstrate that our inverse image-based rendering works well with various challenging datasets without any retraining or finetuning after once trained on synthetic dataset, and outperforms relevant state-of-the-art novel view synthesis methods.
\end{abstract}

\vspace{-2mm}
\section{Introduction}
\label{sec:intro}
\vspace{-1mm}

The selective control of focus and shallow depth of field (DoF) have been critical tools of photography.
  Unfortunately, modern devices such as cell phones have struggled to reproduce these effects because of their small sensors and lenses.
  As a solution to this issue, a concept of 4D light fields \cite{adelson1992single,ng2005light,ng2005fourier}, taking colors and directions of the light flow in a space, enables to render novel views and photographic effects such as refocusing and shallow depth of field.
  However, capturing real light fields requires specialized cameras, and suffers from an inherent trade-offs between spatial and angular resolutions of captured images because one sensor should take both of them. The inherent trade-off potentially causes aliasing when we implement the photographic effects from fewer angular resolutions. Light field angular super-resolutions~\cite{levin2010linear,jin2020learning,jin2020deep,wu2021revisiting,gao2023spatial} have been proposed to mitigate this trade-off, but still need geometrically well-aligned multiple images as input.

\fontsize{10}{11.5}\selectfont
\begin{figure}[t]
\centering
\includegraphics[width=\columnwidth,clip]{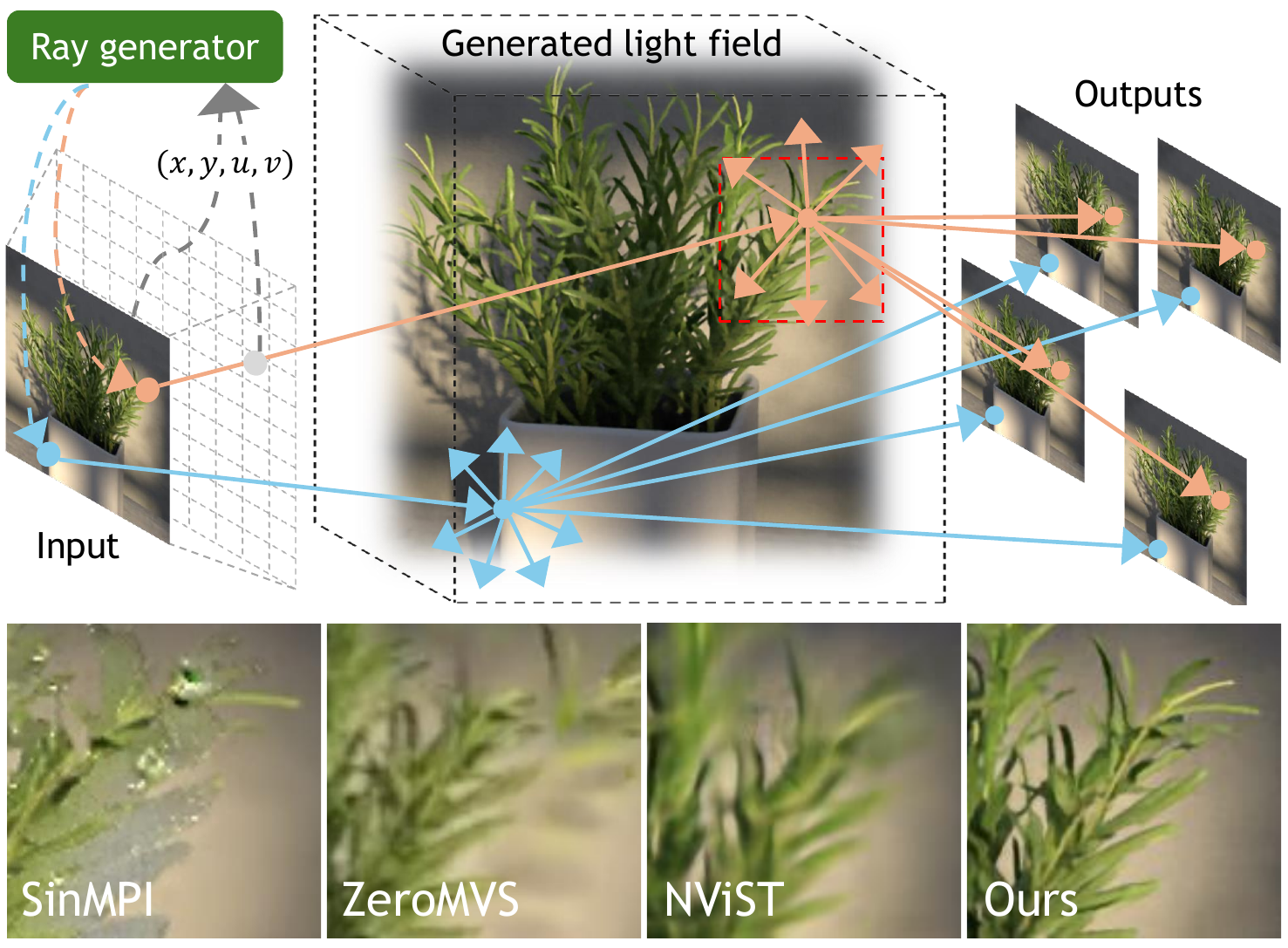}
\vspace{-7mm}
\caption{iIBR enables to generate high fidelity and consistent 4D light field from single image}
\label{fig:teaser}
\vspace{-5mm}
\end{figure}

  Recent advancements in learning-based methods for novel view synthesis allow us to synthesize angular contents of light field. Techniques like NeRF~\cite{mildenhall2020nerf} and 3D Gaussian Splatting~\cite{kerbl20233d} facilitate the transformation of photographs of real-world scenes into 3D models by optimizing the underlying geometry and visual properties. However, producing highly detailed scenes is still a demanding task that requires capturing a large number of images. Its inadequate observations can result in models with incorrect geometry and appearance, leading to unrealistic renderings from novel viewpoints. 
  
  Fortunately, with the recent development of an end-to-end depth-aware view synthesis~\cite{liu2018geometry}, neural rendering~\cite{xu2022sinnerf}, scene approximation into multiple depth planes~\cite{tucker2020single, han2022single, li2020synthesizing} and image generation~\cite{wu2024reconfusion}, we can synthesize images with novel viewpoints from single images to reduce the dependency on dense multi-view captures. In spite of this, they have their own limitations related to quality, efficiency and generality. One of common issues on rendering quality often stems from misaligned geometry and correspondences, particularly in novel view generation approaches. These misalignments in unseen contents of the target viewpoint frequently produce blurry artifacts on 3D objects.


\fontsize{10}{11.5}\selectfont
\begin{figure}[t]
\centering
\includegraphics[width=\columnwidth,clip]{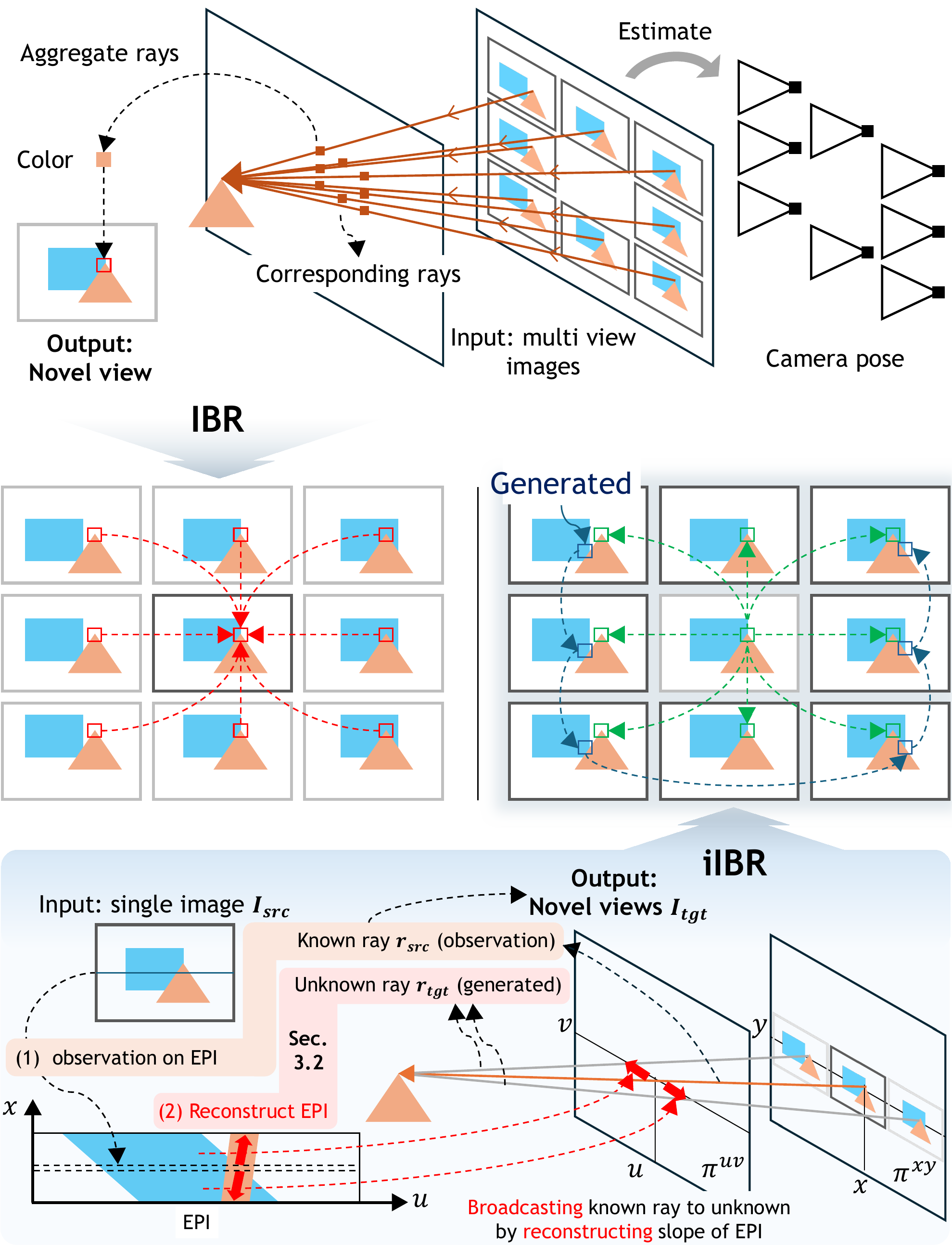}
\vspace{-7mm}
\caption{\textbf{Difference between IBR and iIBR.} IBR renders target image from multiple source image by aggregating correspondence colors. On the contrary, iIBR renders multiple target images from single image by reconstructing unknown light flows, which is shown as lines in the EPI.
EPI is 2D slice of 4D light field. A straight line in EPI represents set of correspondence's light flows sampled form different angular resolution.}
\label{fig:ray_embedding}
\vspace{-5mm}
\end{figure}

  In this paper, we focus on correspondence alignment among generated novel views for better photographic reproduction of scenes. 
  Misaligned pixels along an angular axis of the light field can lead to unwanted artifacts when rendering photographic effects.
  Our key idea comes from a concept of epipolar plane images (EPI)~\cite{bolles1987epipolar} from two-plane parameterized light field~\cite{levoy1996light}. EPI is 2D slices of constant angular and spatial directions in a 4D space. It can be viewed as a 2D image, with spatial resolution along a horizontal axis and angular resolution along a vertical axis. In an EPI, line structures are visible, and their slopes vary based on the disparity among sub-aperture images, whose example is described in \Fref{fig:ray_embedding}. Pixels along a slope are correspondences between sub-aperture images placed in either one column or row on a regular grid. Therefore, each line in the EPI represents a set of light flows of the rays cast from corresponding pixels. The bottom-sided illustration in \Fref{fig:ray_embedding} shows that this EPI's property can be leveraged to generate unknown rays of correspondences from known ray of input image.

  To do this, we formulate light field generation from single images as an inverse problem of image-based rendering which typically synthesize a single image by blending colors from multiple correspondences across different views.
  We propose \textbf{iIBRnet}, \textit{\textbf{i}nverse \textbf{I}mage-\textbf{B}ased \textbf{R}endering} network, a neural renderer that takes single images as input to reconstruct continuous signals of light flows in space, which is the ultimate goal of classical 4D light field imaging, to generate novel views. The concepts of IBR and iIBR are depicted in \Fref{fig:ray_embedding}.
  To render the light flows to novel views, we utilize the Transformer~\cite{vaswani2017attention} to compute self-attention scores of ray embeddings from the input image and the target novel viewpoint.
  This procedure reconstructs angular-consistent, high-fidelity light field images. Additionally, we improve the generality of iIBRnet through pixel-level processing for novel view synthesis. Our model is trained on only synthetic images and tested on real-world images without any re-training or fine-tuning. We demonstrate that our method produces state-of-the-art results in light field generation compared to relevant works, showcasing notable generalization performance.

\vspace{-1mm}
\section{Related Works}
\vspace{-1mm}

\subsection{Image-based rendering}

  Image-based rendering (IBR)~\cite{shum2000review} has emerged with desire on making free-veiwpoint images, given multiple images. It enables the synthesis of novel views from collection of input images. The light field~\cite{levoy1996light} allows us to parameterize incoming light flows from world coordinates to describe scene structures, which are formulated as 4D plenoptic function~\cite{isaksen2000dynamically}. However, light field rendering requires a dense sampling of input views to yield high quality images. To mitigate the dense sampling constraint, Lumigraph~\cite{gortler2023lumigraph} uses an approximate geometry.
  Co-operating with explicit geometry~\cite{choi2019extreme, riegler2021stable} shows plausible rendering quality with few image samples. However, learning 3D proxy geometry is challenging, and errors in this process can result in misaligned correspondences during rendering.

  Recently, the concept of IBR has contributed to learning radiance fields by aggregating corresponding visual features. Methods for finding correspondences can be categorized as: aligning along an epipolar line and aggregating their colors while leveraging the capabilities of transformers~\cite{suhail2022generalizable, charatan2024pixelsplat}; collecting visual features from input images during volumetric sampling~\cite{yu2021pixelnerf, wang2021ibrnet}; and using plane sweep volumes~\cite{chen2024mvsplat}.
  They achieve generality, enabling the representation of scenes with only a forward-pass. 
  However, they requires multiple input images and precise camera poses, while our method requires only single images.

\subsection{Novel view synthesis from single images}

  Novel view synthesis from single images is challenging due to a ill-posed nature on representing scene geometry. Previous methods obtain geometric information by using either single-image depth estimation~\cite{srinivasan2017learning, liu2018geometry}
  or mesh estimation~\cite{hu2021worldsheet}. With the estimated depth information, works in \cite{han2022single, pu2023sinmpi} compute multi-plane images (MPIs) to approximately account for scene geometry, which can be projected onto novel viewpoints. 3D photo conversions from single images~\cite{shih20203d} separate foreground and background of scenes, using soft occlusion masks to fill missing regions with plausible contents through inpainting. SinNeRF~\cite{xu2022sinnerf} and NViST~\cite{jang2024nvist} infer radiance fields for scenes from only single images. However, these methods heavily rely on large-scale datasets, as their networks focus on leveraging useful intuitions about scene information through the visual feature extraction. Our method overcomes this issue by treating pixels as individual light flows.

  Given that text-to-image generation models are highly effective at producing visually promising images, methods for obtaining multi-view observations from single images have been proposed~\cite{liu2023zero, tang2024lgm, wu2024reconfusion}. These models offer stronger priors for unseen contexts of scenes using the input images with pose conditioning. However, because they generate views independently, remaining uncertainties among correspondences can lead to performance drops. Multidiff~\cite{muller2024multidiff} generates novel views by warping true colors from the input image using its corresponding depth map and fills in empty spaces through generation. While the warped true-color pixels are geometrically consistent, the generated regions are not. Our method is free from this problem because we iteratively update inpainted contents at other novel views, ensuring consistent view generation. 

   \fontsize{10}{11.5}\selectfont
   \begin{figure}[t]
    \centering
     \includegraphics[width=\columnwidth]{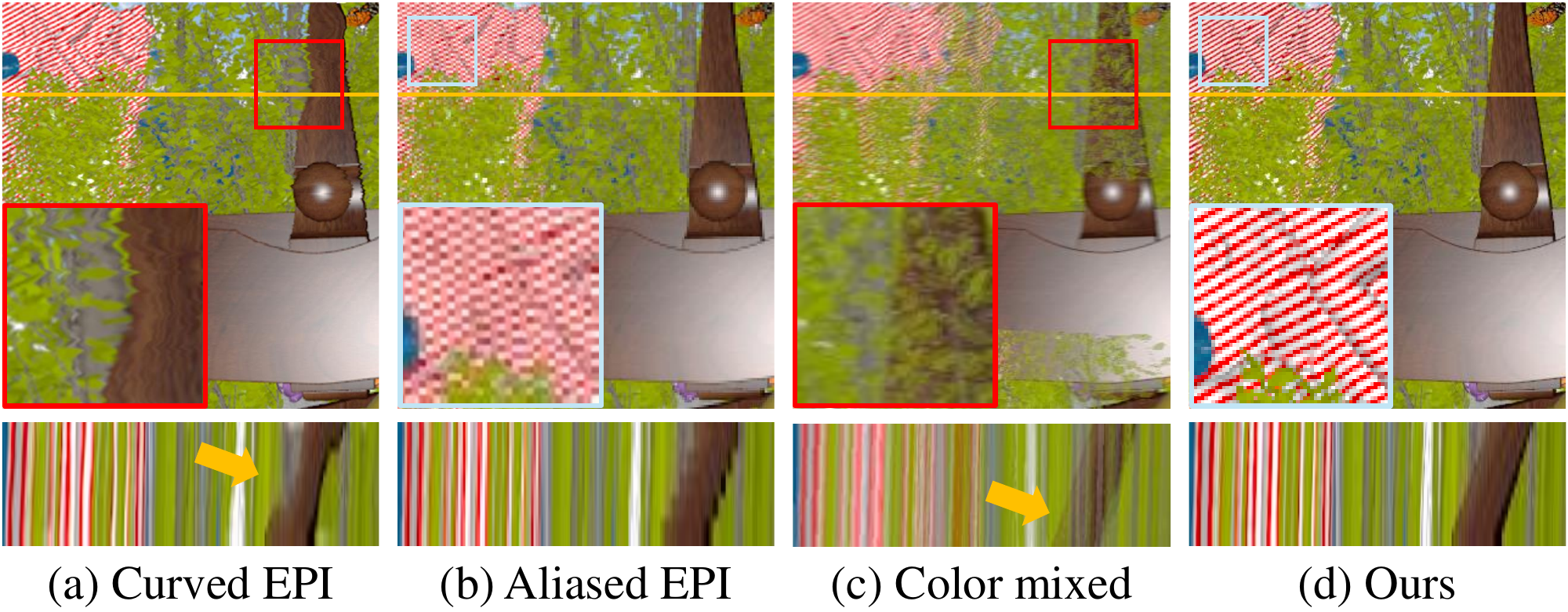}
    \vspace{-7mm}
     \caption{\textbf{Importance of accurate EPI generation.} Restoring accurate 2D rays along an EPI in sub-pixel accuracy is importance for novel view synthesis.}
     \vspace{-5mm}
     \label{fig:bad_epi_to_render}
  \end{figure}

   \fontsize{10}{11.5}\selectfont
   \begin{figure}[t]
    \centering
     \includegraphics[width=\columnwidth]{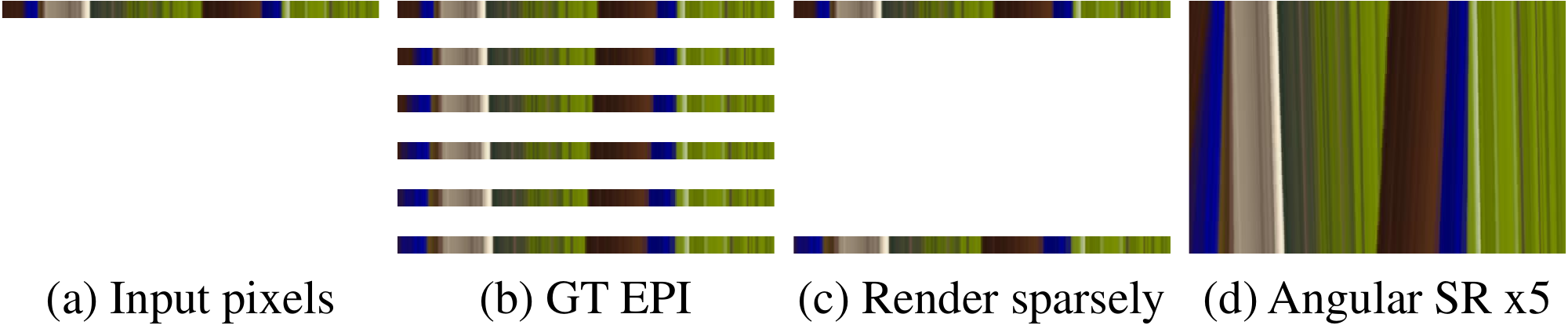}
    \vspace{-7mm}
     \caption{\textbf{2D ray generation of iIBR.} iIBR restores continuous 2D rays from single pixels, as demonstrated by the angular super-resolution of light field images.}
     \vspace{-5mm}
     \label{fig:epi_angular_sr}
  \end{figure}

   \fontsize{10}{11.5}\selectfont
   \begin{figure*}[t]
    \centering
     \includegraphics[width=0.9\textwidth]{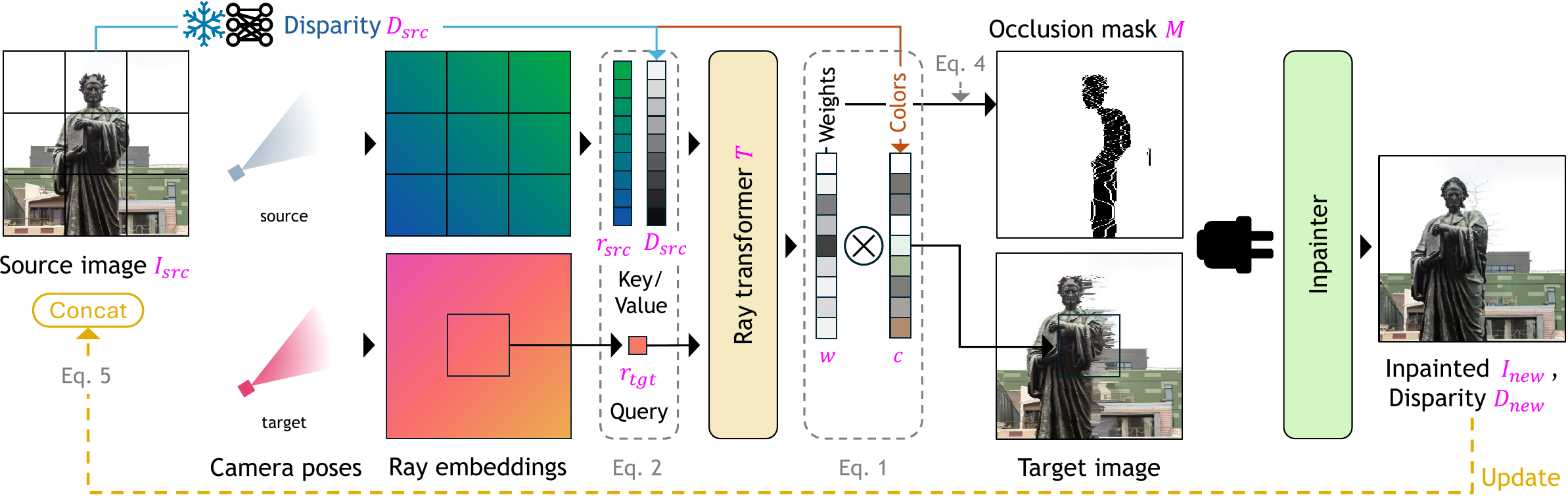}
     \vspace{-3mm}
     \caption{
     \textbf{An overview of iIBRnet.} Given a single image, iIBRnet generates novel views. iIBRnet operates in two key stages: (1) the ray transformer, which calculates the relationships between rays from different viewpoints, acting as the mechanism for rendering the light flows of cast rays from input image, and (2) an inpainting process that handles occlusions and updates the input for iIBRnet.}
     \vspace{-5mm}
     \label{fig:method}
  \end{figure*}

\section{Methodology}\label{sec:method}

Given a single image, our goal is to reconstruct a 4D light field using our iIBR which is implemented as a neural rendering process. \Fref{fig:method} provides an overview of our neural renderer, iIBRnet.
We first define a concept of iIBR and provide technical insights how to incorporate it into a neural rendering network. We then describe an architecture and a rendering pipeline, including occlusion detection and handling. To better explain our method, we start with a 2D case of iIBR, involving 1D spatial and 1D angular dimensions, and then extend it into the 4D light field representation.

\subsection{Inverse image-based rendering}
\label{sec:iibr}

An inverse rendering~\cite{marschner1998inverse, sato1997object} typically refers to a process of reversing physically-based rendering, aiming to estimate physical attributes of a scene—such as geometry, material properties and lighting—from images.
The concept of iIBR starts from an imagination of an ideal IBR. As illustrated in the upper-sided \Fref{fig:ray_embedding}, the ideal IBR would be possible if exact correspondences are available. Since the ideal IBR is theoretically achievable, its inverse problem allows us to propagate colors from the input image to other sub-aperture images at precise locations of each correspondence.

Physically, each pixel in the input images represents a ray of light flows carrying the pixel color. A pixel can be cast into a structured 4D ray space defined by two planes $\pi^{xy}$ and $\pi^{uv}$, where the local plane coordinates at the intersections are  $(x,y) \in \pi^{xy}$ and $(u,v) \in \pi^{uv}$. Consequently, the set of correspondences for the ray $(x,y,u,v)$ from horizontally aligned sub-aperture images can be defined as $S=\{(x_i, y, u_i, v) \mid x_i \neq x, u_i \neq u, \, i \in I \}$, where $I$ is an index set. This point of view simplifies the problem of finding correspondences by reducing it to a task of calculating pairs $(x_i, u_i)$, which functionally serves to obtain the set $S$.
Here, one of our significant contribution is that EPIs are used as a powerful tool for solving this problem because they are constructed along two axes, $x$ and $u$; i.e., constructing accurate EPIs directly addresses the challenge of calculating the set $S$.
Generating EPIs from a single angular content via 2D image processing is highly ill-posed, but representing EPIs by the slopes of pixels could provide a viable solution. The approach involves casting the ray of each input pixel into either $x-u$ or $y-v$ 2D space with the orientation defined by the pixel's slope.


\subsection{2D inverse Image-Based Rendering}

For easy-to-understand our 2D iIBR, we first consider 2D ray space.
The sub-problem of synthesizing a 4D light field via iIBR can be represented in the reduced dimensionality of the ray space. This is achieved by selecting one spatial and one angular domain, thereby considering the 2D ray space, without a loss of generality. 
Our goal is to calculate a set $S$ by casting the ray of an input pixel into 2D ray space, represented as EPIs. Imagine a baby drawing straight lines from top to bottom on a blank page with crayons. These straight lines will form an EPI, with the colors of crayons matching the colors of the input pixels.

Drawing a line in the EPI precisely, i.e., casting a 2D ray in the 2D ray space, is critical for synthesizing accurate correspondences from the input image to sub-aperture images.
Challenges arise when the drawn line is not straight (i.e. curved), or when the line needs to be drawn at sub-pixel coordinates. These lead to geometrically inconsistent results and aliasing, as illustrated in \Fref{fig:bad_epi_to_render} (a) and (b), respectively. 

To address this, our neural renderer, iIBRnet, is designed to render each angular content step-by-step, progressing from the input to the next angular content and so on. It renders a pixel by a weighted summation of the colors obtained through tracking the surrounding 2D rays. With this consideration, iIBRnet is capable of generating continuous 2D rays with sub-pixel accuracy, enabling anti-aliased rendering and unlimited angular super-resolution. It could also render views sparsely, where the step of angular content generation can be skipped, as demonstrated in \Fref{fig:epi_angular_sr}.


Rendering a color of 2D ray ${c}_{i}^{j}$ in the $i$-th spatial and $j$-th angular dimension is achieved by aggregating all spatial contents in the $(j-1)$-th angular dimension, each associated with their respective weights $w$, as shown below:
\begin{equation}
{
 \mathbf{c}_{i}^{j} = \sum_{a \in A_{j-1}} w_{a}^{j-1} c_{a}^{j-1}
 ,
}
\label{eq:render}
\end{equation}
where $A_{j-1}$ is a set of spatial indices of pixels in the $(j-1)$-th angular dimension. iIBRnet is designed to predict the weight $w$ rather than directly predicting the color ${c}_{i}^{j}$.

In the context of aggregation, we use Transformer architecture~\cite{vaswani2017attention} to predict $w$. Transformers are widely used in neural rendering~\cite{wang2021ibrnet, suhail2022generalizable} due to their effectiveness in aggregating visual information. However, to be more precise, our iIBRnet focuses on investigating the relationships between virtually projected rays rather than feature aggregation. In physical terms, predicting $w$ involves establishing connections between each 2D ray in the set $A_{j-1}$ and the 2D ray associated with ${c}_{i}^{j}$, and determining how closely they are related. Therefore, our ray transformer in iIBRnet uses only ray coordinates as inputs, without any visual feature. The ray coordinates are sampled from a camera of the input image (typically with the view matrix defined as an identity matrix) and from the camera positions of the novel views to be rendered.

The representation of rays is also essential to predict $w$. Since we assume a 4D ray space with a two-plane parameterized light flow, the ray coordinates are defined as a light slab~\cite{levoy1996light}, denoted by $(x,y,u,v)$. In order to generalize new scenes without relying on specific camera configurations, we choose to parameterize rays using Plücker coordinates which has been used to model a neural field~\cite{sitzmann2021light}.
Plücker coordinates represent a ray that originates from a point $o \in \mathbb{R}^3$ and casts in the direction $d \in \mathbb{R}^3$ as $r=(d, o \times d)$. This representation spans four degrees of freedom and two scale factors within six dimensions, allowing us to uniquely process and define rays.

Another benefit of leveraging angular information in an EPI is to provide geometric information. Its angle of each slope directly represent disparity, while the slopes are formed due to the uniform sampling of corresponding light flows. From a perspective of an inverse problem, geometric priors help guide the correct formation of EPI slopes and resolve the challenge of distinguishing foreground and background pixels in regions where pixel slopes intersect. To incorporate such geometric priors, we focus on the positional encoding of the ray transformer. 
In GPNR~\cite{suhail2022generalizable}, a novel positional encoding is proposed for the Transformer architecture to retain the spatial position of visual information, epipolar geometry and relative camera positions. Similarly, we introduce a positional encoding to embed them on the matching direction of 2D rays.
This gives us an insight how the ray transformer in iIBRnet functions: it learns to establish a set of potentially matched correspondences based on the 2D ray direction. We choose disparity information for the positional encoding because disparity values have broader purposes than just representing scene depths. In the EPI, disparity values represent the displacement along the $u$-axis for each unit of movement along the $x$-axis, effectively capturing the 2D ray direction.
The disparity is directly inferred from a depth foundation model $\mathcal{F}$ (we use DepthAnythingv2~\cite{depth_anything_v2} in this work), and is further refined using scale factors $\alpha$ and shift factors $\beta$ predicted by a simple convolutional neural network. 
The final disparity value $D$ is calculated as: $D_{src} = \alpha\mathcal{F}(I_{src})+\beta$, where $I_{src}$ is the input image. Note that any depth estimation model can easily be available in our framework, and recent metric depth estimation~\cite{piccinelli2024unidepth} may further minimize the need for scale/shift parameters.
Finally, our $w$ prediction using the ray transformer can be formulated as follows:
\begin{equation}
{
 w^j = T(\{\,[\, r_{tgt,a}^j \mid\mid r_{src,a}^j \mid\mid D_{src,a}^j \,] \mid a \in A_j \,\})
 ,
}\end{equation}
where $T$ refers to the ray transformer, and $r_x^j$ and $d_x^j$ denote the 2D slices of $r$ and $d$ at the $j$-th angular dimension, respectively. 

To optimize $T$, the objective function of iIBRnet consists of three loss terms: The first is $L_2$ loss on the rendered color, $\mathcal{L}_c = || c - \hat{c} ||^2_2$ where $\hat{c}$ is its ground truth color. The second term is an entropy loss on $w$, $\mathcal{L}_w = -\sum w log(w)$, ensuring that a dominant 2D ray contributes—though this may not exist in occluded regions— to rendering the target 2D ray. This helps prevent the issue on blending background and foreground contents, as illustrated in \Fref{fig:bad_epi_to_render} (c). The last term is $L_1$ loss on the local structure tensor~\cite{wanner2012globally} of the rendered EPI $	\zeta$, $\mathcal{L}_{epi} = || J(\zeta) - J(\hat{\zeta}) ||$, where $J$ and $\hat{\zeta}$ denote the structure tensor operator and ground truth EPI, respectively. This loss encourages the rendering of 2D rays to have straight linear structures on EPIs, and assists in predicting the scale/shift value of disparity by refining the local slopes of the rendered EPIs. In total, our objective function is formulated as follows:
\begin{equation}
{
 \mathcal{L} = \lambda_c\mathcal{L}_c + \lambda_w\mathcal{L}_w + \lambda_{epi}\mathcal{L}_{epi}
 .
}\end{equation}

   \fontsize{10}{11.5}\selectfont
   \begin{figure}[t]
    \centering
     \includegraphics[width=\columnwidth]{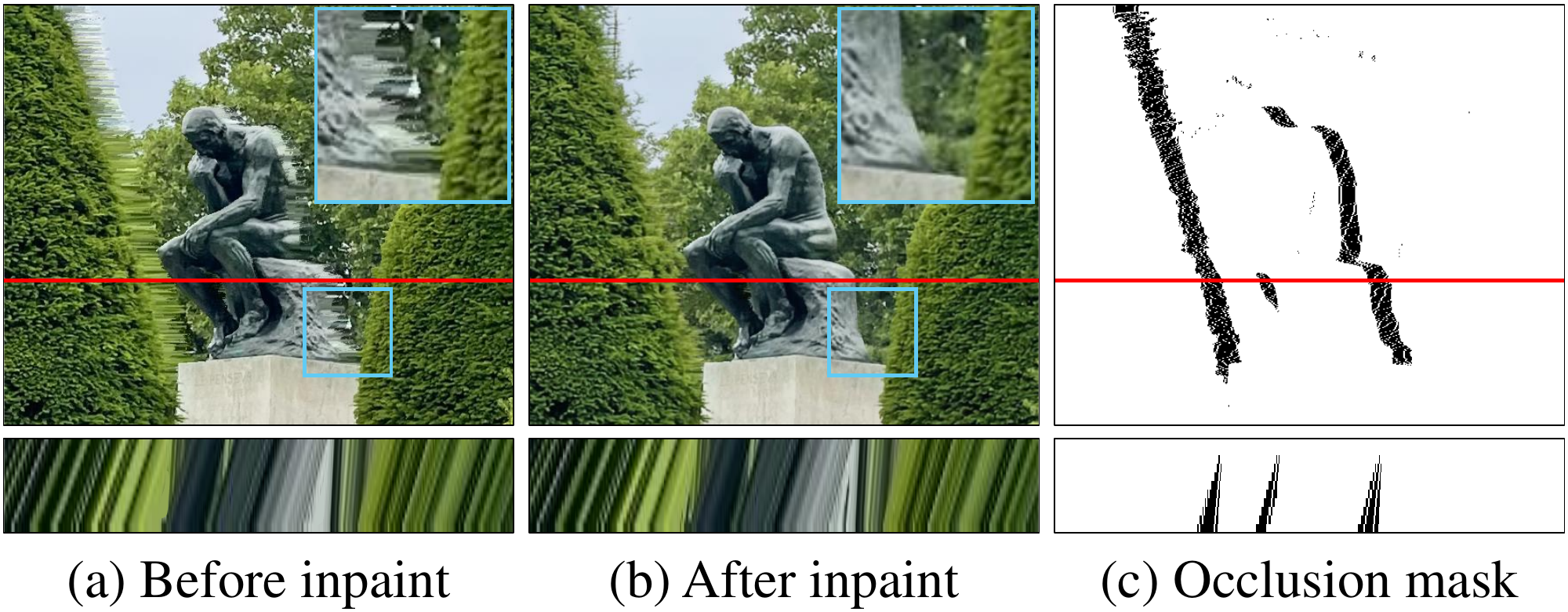}
     \vspace{-7mm}
     \caption{\textbf{Occlusion handling on 2D iIBR.} We can effectively capture occluded regions. The inpainted background is then regenerated to form 2D rays through the 2D iIBR.}
     \vspace{-5mm}
     \label{fig:epi_pred_eg}
  \end{figure}

\subsection{Occlusion handling}

While our ray transformer is capable of accurately rendering 2D rays, challenges remain in restoring contents in occluded regions.
To address this issue, we need to generate unseen contexts in the occluded regions. Thanks to recent advances in recent generative models~\cite{rombach2022high, saharia2022palette, yu2019free}, inpainting occlusions, seamlessly matching the surrounding context, is feasible.

As the first step, we detect occlusions in the synthesized novel view. In \Fref{fig:epi_pred_eg}, we present an example of an EPI. One thing to note is that in the initially rendered EPI from the ray transformer, occlusions appear as a blend of adjacent foreground and background contents. This implies that the weight $w$ used to render pixels in occluded regions may not accurately target the dominant 2D rays corresponding to the occlusion contents. Therefore, we interpret $w$ as an uncertainty in rendering and detect an occlusion mask $M$ using the following formulation:
\begin{equation}
{
M_{i}^{j} = \begin{cases} 
	0 & \text{if } -\sum_{a \in A_{j-1}} w_{a}^{j-1} log(w_{a}^{j-1}) < k \\ 
        1 & \text{else} 
    \end{cases},
}\end{equation}
where $k$ is a threshold, and empirically set to 2.3. For occlusions, $M$ is 0. We then restore the occluded content by inpainting the image from a novel viewpoint for the masked regions. However, since 2D iIBR renders only a single slice of a 2D image at a time, we render the complete 2D novel view image at first, and then inpaint the occlusions. We use Latent diffusion model~\cite{rombach2022high} for inpainting.

Here, we aim to achieve consistent inpainting across all sub-aperture images. After the complete novel view is generated through inpainting, we treat each newly generated pixel as a new ray, allowing us to cast it into the 2D ray space using 2D iIBR. 
We first infer disprity as $D_{new} = \alpha\mathcal{F}(I_{new})+\beta$, where $I_{new}$ is the novel view. We then assign ray coordinates and the corresponding disparity to the generated pixels from the masked region, and incorporate them into the input set $ \{r_{src,x}^j, D_{src,x}^j \mid x \in A_j\}$ for the ray transformer. This process is iteratively repeated whenever we encounter an occlusion that needs to be generated.

   \fontsize{10}{11.5}\selectfont
   \begin{figure*}[t]
    \centering
     \includegraphics[width=\textwidth]{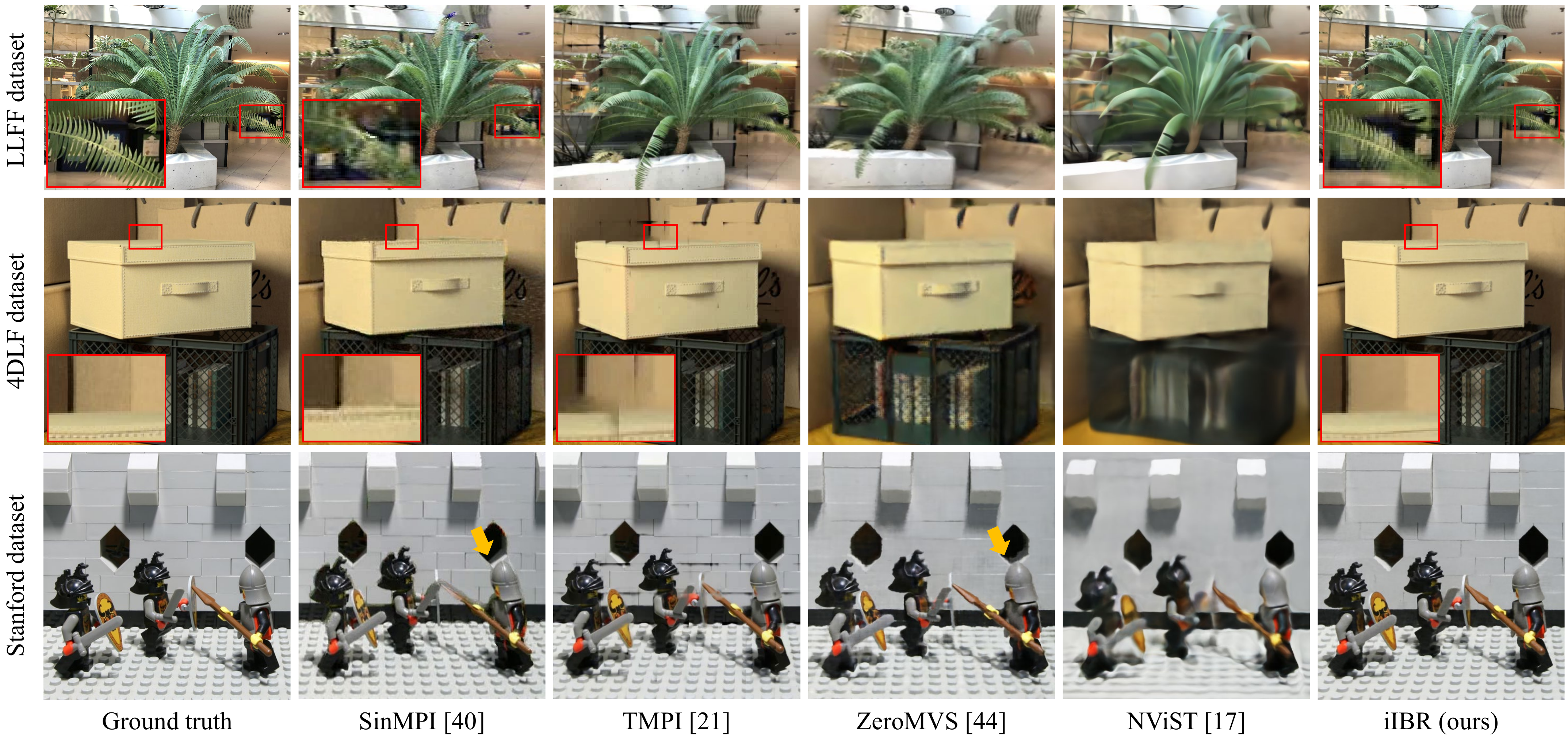}
     \vspace{-7.5mm}
     \caption{\textbf{Qualitative comparison.} Our iIBR consistently synthesizes novel views, an archives highest quality.}
     \vspace{-2mm}
    \label{fig:qualitative_results}
  \end{figure*}

\subsection{4D light field generation}

With the solution to the sub-problem of 2D iIBR, it is straightforward to extend this into (2+1)D iIBR, where one spatial dimension is expanded. The process involves simply repeating the 2D iIBR steps in the extended domain. For instance, if we apply iIBR into a 2D image along a single angular dimension, a viable solution would be to perform 2D iIBR on each horizontal slice of the input image, and then merge all the slices.

For a 4D ray space where includes an additional extension in the angular dimension, we opt to expand the positional encoding accordingly. Since the input ray coordinate is already defined in the 4D ray space, the ray transformer for 4D iIBR is formulated as follows:
\begin{equation}
\begin{split}
 w^{j_1, j_2} = T(\{[\, r_{tgt,a,b}^{j_1, j_2} \mid\mid r_{src,a,b}^{j_1, j_2} \mid\mid D_{x,src,a,b}^{j_1, j_2}  \mid\mid  D_{y,src,a,b}^{j_1, j_2} \,]  \\
 \mid a \in A^{j_1}, \, b \in A^{j_2} \}), \quad \text{(5)} \nonumber
\end{split}
\label{eq:4Dweight}
\end{equation}
where $D_x$ and $D_y$ denote the disparity between sub-aperture images along the horizontal and vertical angular axes, respectively. After rendering through the ray transformer, inpainting occlusions and iterative update for the set of the input 4D ray are also performed. To reduce the number of iterations and enhance the efficiency, we first generate the farthest views from the center view, and then update the generated occlusion rays.

   \fontsize{10}{11.5}\selectfont
   \begin{figure*}[t]
    \centering
     \includegraphics[width=\textwidth]{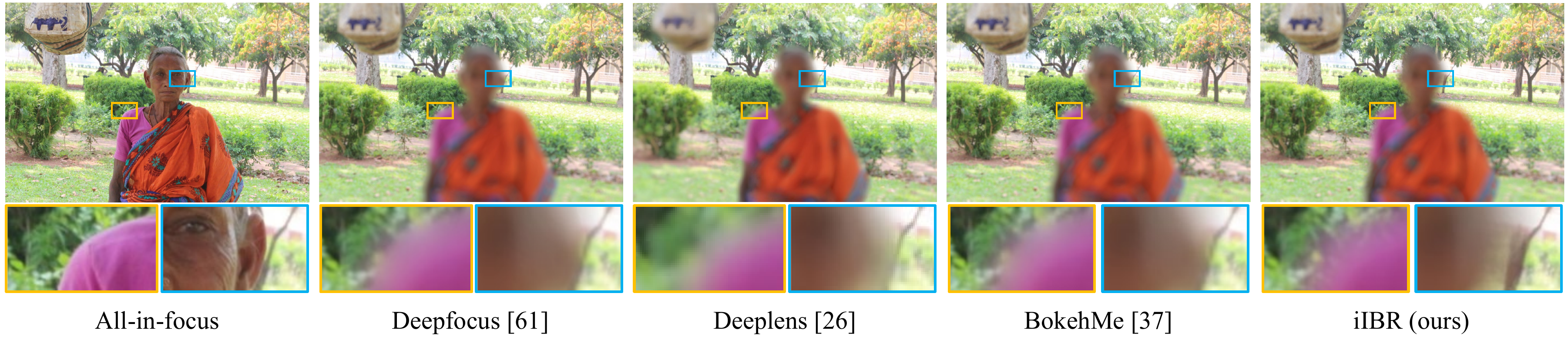}
     \vspace{-7mm}
     \caption{\textbf{Qualitative comparison for digital refocusing.} Since iIBR generates 4D light flows in space and rendering light field image, physical and realistic digital refocusing is available.}
     \vspace{-2mm}
     \label{fig:refocus}
  \end{figure*}

\section{Experiments}\label{sec:experiment}
Further evaluations and analyses are provided in our \textbf{supplementary material}, which includes: (1) \textit{Video results} demonstrating the qualitative results of our novel view synthesis across various scenes; (2) \textit{BRDF rendering} details, explaining how our learned iIBRnet is used for rendering specularities; and (3) additional experimental results and analysis on our rendering pipeline.

\subsection{Implementation details}

  We implement our network using a public Pytorch~\cite{NEURIPS2019_9015} framework.
  For training, we use Adam~\cite{kingma2014adam} optimizer with $\beta_1=0.9$ and $\beta_2=0.99$. The learning rate and batch size are set to $0.0001$ and 1, respectively.
  Our network is trained on a single NVIDIA Tesla V100 GPU for a day, 630K iterations. To avoid an overfitting problem, we adopt a data augmentation in~\cite{heber2017neural}. 
  The balance terms for the loss functions are set to $\lambda_c=100$, $\lambda_w=1$, and $\lambda_epi=0.1$.
  For memory-efficient training on GPU, we select five source ray coordinates closest to the target ray coordinate to learn $T$.

  \begin{table*}[t] 
    \centering
    \fontsize{8}{9}\selectfont
    \resizebox{\textwidth}{!}{
    \begin{tabular}{@{}l@{}c@{}c@{}c@{}c@{}c@{}c@{}c@{}c@{}c@{}c@{}c@{}c@{}c}
    \toprule
    \multicolumn{2}{c}{}&\multicolumn{3}{c}{Pov-ray dataset}& \multicolumn{3}{c}{4DLF dataset}& \multicolumn{3}{c}{Stanford dataset}& \multicolumn{3}{c}{NeRF LLFF dataset} \\
    \cmidrule(r){3-5}
    \cmidrule(r){6-8}
    \cmidrule(r){9-11}
    \cmidrule(r){12-14}
    (ft:fine-tuning) & ~~~ & ~~~~PSNR$\uparrow$~~~ & ~~~SSIM$\uparrow$~~~ & ~~~~LPIPS$\downarrow$~~~ & ~~~~PSNR$\uparrow$~~~ & ~~~SSIM$\uparrow$~~~ & ~~~~LPIPS$\downarrow$~~~ & ~~~~PSNR$\uparrow$~~~ & ~~~SSIM$\uparrow$~~~ & ~~~~LPIPS$\downarrow$~~~ & ~~~~PSNR$\uparrow$~~~ & ~~~SSIM$\uparrow$~~~ & ~~~~LPIPS$\downarrow$~~~
    \\
    \midrule
    SinMPI~\cite{pu2023sinmpi}&
    & 22.419 & 0.697 & 0.213
    & 22.432 & 0.701 & 0.201
    & 22.381 & 0.674 & 0.218
    & 17.101 & 0.530 & 0.581
    \\
    TMPI~\cite{khan2023tiled}&
    & 23.619 & 0.726 & 0.170
    & 24.832 & 0.780 & 0.132
    & 24.013 & 0.749 & 0.162
    & 17.561 & 0.569 & 0.422
    \\
    ZeroMVS~\cite{liu2023zero}&
    & 20.835 & 0.710 & 0.229
    & 20.759 & 0.689 & 0.231
    & 20.481 & 0.664 & 0.228
    & 11.211 & 0.403 & 0.623
    \\
    NViST~\cite{jang2024nvist}&
    & 19.889 & 0.553 & 0.290
    & 19.842 & 0.540 & 0.325
    & 18.548 & 0.471 & 0.493
    & 15.341 & 0.437 & 0.701
    \\
    \midrule
    iIBR (ours) (zero-shot)~~~~~~~ &
    & \textbf{28.407} & \textbf{0.931 }& \textbf{0.037}
    & 28.095 & 0.926 & 0.046
    & \textbf{27.889} & \textbf{0.910 }& 0.068
    & 24.910 & 0.810 & 0.183
    \\
    iIBR (ours) (ft) &
    & \textbf{28.407} & \textbf{0.931 }& \textbf{0.037}
    & \textbf{28.382} & \textbf{0.929 }& \textbf{0.041}
    & 27.682 & 0.902 & \textbf{0.052}
    & \textbf{25.534} & \textbf{0.883 }& \textbf{0.095}
    \\
    \bottomrule
    \\
    \end{tabular}
    }
    \vspace{-5mm}
    \caption{\textbf{Qualitative comparisons.} iIBR outperforms all compared methods across datasets and metrics. We tested iIBR in a zero-shot setting using the model pretrained on the Pov-ray dataset, as well as fine-tuned versions for each specific dataset.}
    \vspace{-4mm}
    \label{tab:quant}
  \end{table*}

  \begin{table}[t] 
    \centering
    \fontsize{8}{9}\selectfont
    \resizebox{\columnwidth}{!}{
    \begin{tabular}{@{}l@{}c@{}c@{}c@{}c@{}c@{}c@{}c@{}c@{}c@{}c}
    \toprule
    \multicolumn{2}{c}{Test set $\rightarrow$}&\multicolumn{3}{c}{Pov-ray}& \multicolumn{3}{c}{4DLF}& \multicolumn{3}{c}{Stanford} \\
    \cmidrule(r){3-5}
    \cmidrule(r){6-8}
    \cmidrule(r){9-11}
    Train set $\downarrow$ & ~~~ & ~~~PSNR$\uparrow$~~ & ~~SSIM$\uparrow$~~ & ~~LPIPS$\downarrow$~~ & ~~~PSNR$\uparrow$~~ & ~~SSIM$\uparrow$~~ & ~~LPIPS$\downarrow$~~ & ~~~PSNR$\uparrow$~~ & ~~SSIM$\uparrow$~~ & ~~LPIPS$\downarrow$~~
    \\
    \midrule
    Pov-rayz
    & ~ & 28.407 & 0.931 & 0.037
    & 28.095 & 0.926 & 0.046
    & 27.889 & 0.910 & 0.068
    \\
    4DLF
    & ~ & 28.183 & 0.926 & 0.048
    & 28.382 & 0.929 & 0.041
    & 27.691 & 0.903 & 0.068
    \\
    Stanford
    & ~ & 27.301 & 0.901 & 0.069
    & 27.163 & 0.892 & 0.069
    & 27.682 & 0.902 & 0.052
    \\
    \bottomrule
    \\
    \end{tabular}
    }
    \vspace{-5mm}
    \caption{\textbf{Zero-shot cross validation.} iIBR demonstrates consistent performance across various datasets, showing high generalizability for novel view synthesis.}
    \vspace{-3mm}
    \label{tab:zeroshot}
  \end{table}

  \begin{table}[t] 
    \centering
    \fontsize{8}{9}\selectfont
    \resizebox{\columnwidth}{!}{
    \begin{tabular}{@{}l@{}c@{}c@{}c@{}c}
    \toprule
    ~~~~~~~~~~~~ & ~~~~~~ & ~~~~~~PSNR$\uparrow$~~~~~ & ~~~~~SSIM$\uparrow$~~~~~ & ~~~~~LPIPS$\downarrow$~~~~~
    \\
    \midrule
    w/o disparity positional encoding
    &  & 12.008 & 0.312 & 0.808
    \\
    w/o inpainting
    &  & 24.899 & 0.735 & 0.163
    \\
    w/o entropy loss
    &  & 25.210 & 0.831 & 0.139
    \\
    w/o EPI structure tensor loss
    &  & 26.910 & 0.882 & 0.116
    \\
    \midrule
    replace depth model to MiDaS~\cite{ranftl2020towards}
    &  & 27.990 & 0.857 & 0.063
    \\
    replace depth model to ZoeDepth~\cite{bhat2023zoedepth}
    &  & 28.310 & 0.901 & 0.043
    \\
    replace depth model to UniDepth~\cite{piccinelli2024unidepth}
    &  & 28.372 & 0.930 & 0.040
    \\
    \midrule
    Ours
    &  & \textbf{28.407} & \textbf{0.931} & \textbf{0.037}
    \\
    \bottomrule
    \\
    \end{tabular}
    }
    \vspace{-5mm}
    \caption{\textbf{Ablation study.} An ablation study was conducted on the Pov-ray dataset, demonstrating that each key component of our method contributes to the performance of iIBR.}
    \vspace{-3mm}
    \label{tab:ablation}
  \end{table}

\noindent\textbf{Dataset.}\quad
We evaluate ours and state-of-the-art methods on four datasets: three light field datasets used for training and one real-world dataset to assess real-world performance.

\noindent(1) Pov-ray dataset~\cite{heber2016convolutional}: Pov-ray dataset is a synthetic light field dataset with $11\times11$ sub-aperture images on a regular grid. The dataset contains 900 scenes, and we divide them into 800 training scenes and 100 test scenes, following the authors' split.

\noindent(2) Stanford dataset~\cite{stanfordlightfield}. We use the (new) Stanford Light Field Archive dataset, which was captured in a camera array. This dataset has $17\times17$ sub-aperture images for each scene. The dataset is primarily used for evaluation, but we also report fine-tuned results on this dataset.

\noindent(3) 4D light field dataset (4DLF)~\cite{honauer2016benchmark}: The synthetic dataset consists of $9\times9$ sub-aperture images. Similar with the Stanford dataset, this dataset is mainly utilized for evaluation and fine-tuned results.

\noindent(4) NeRF LLFF dataset~\cite{mildenhall2019local}: This dataset provides unstructured multiview images captured in real-world scenarios. To evaluate our 4D iIBR on this dataset, we first densely construct a structured 4D light field(128$\times$128) from the view closest to a center of all cameras. We then render free views for evaluation by utilizing an approach from~\cite{isaksen2000dynamically}.
This dataset is mainly utilized for evaluation and fine-tuned results as well. To fine-tune the dataset to our method, we use a coarse 4D light field rendered through ZipNeRF~\cite{barron2023zipnerf}.

\noindent\textbf{Evaluation protocol.}\quad
We compare our 4D iIBR with reproducible state-of-the-art methods that are capable of synthesizing full scenes, not just object-centric scenes.
For a fair comparison, all the compared methods, including ours, use the same resolution of input images and depth maps from DepthAnythinv2~\cite{depth_anything_v2}. Since SinMPI~\cite{pu2023sinmpi} utilizes an off-the-shelf latent diffusion model checkpoint~\cite{rombach2022high} , therefore we apply our inpainter fine-tuned with each dataset. 

To evaluate performances, we use common quantitative measures of the image quality: peak signal-to-noise ratio (PSNR), structural similarity index measure (SSIM)~\cite{wang2004image} and learned perceptual image patch similarity (LPIPS)~\cite{zhang2018unreasonable}.

\subsection{Qualitative results}

\noindent\textbf{Novel view synthesis.}\quad
In \Fref{fig:qualitative_results}, we present a qualitative comparison for novel view synthesis. We render top-left sub-aperture image from bottom-left sub-aperture image, while a center view is used for NeRF LLFF dataset. Since our method is designed to accurately reconstruct correspondences for novel view synthesis, it consistently produces precise results, even in areas with fine structures. Although 4D iIBR computes relationships among rays, its individual ray processing enables our model to handle any input image, regardless of scene structures or contexts. Additionally, The better performance of our method on the NeRF LLFF dataset comes from the capability of the dense light field prediction and outpainting.
In contrast, the comparison methods struggle to reconstruct fine details because they rely heavily on visual features. While SinMPI does not heavily depend on visual information, it exhibits an issue on depth scale misalignment when generating out-painted images, even when the depth estimator is fine-tuned for the scene. SinMPI often fails to produce accurate inpainted texture, given a lack of details in the occlusion mask.

\begin{figure}[t]
    \centering
     \includegraphics[width=\columnwidth]{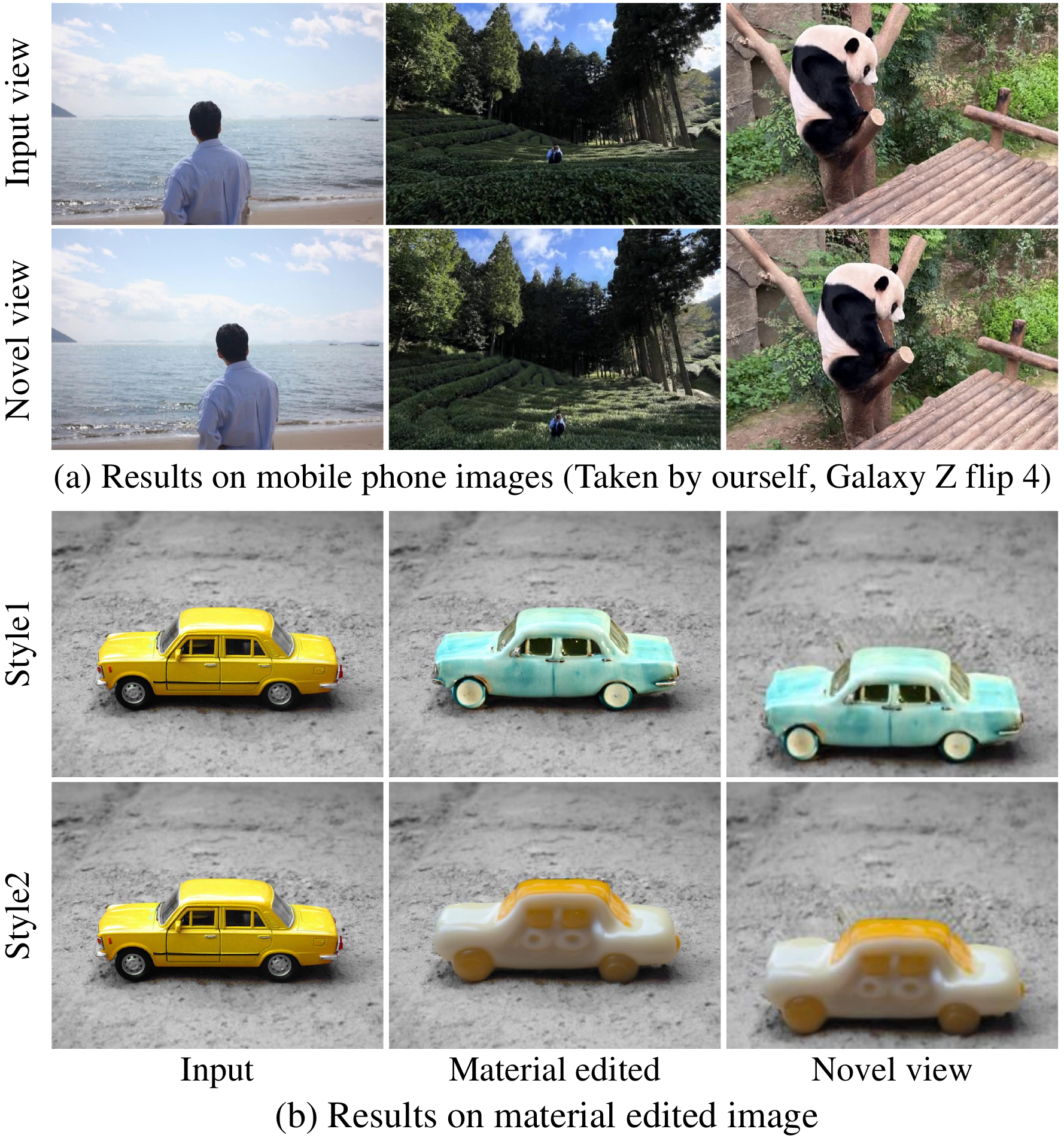}
     \vspace{-7mm}
     \caption{\textbf{Zero-shot novel view synthesis results.} Our iIBR can synthesize novel views with any image and demonstrates consistent performance regardless of the scene context.}
     \vspace{-5mm}
     \label{fig:zeromat}
  \end{figure}

\noindent\textbf{Zero-shot result.}\quad
Our iIBR can synthesize novel views with any image and demonstrates consistent performance regardless of the scene context. In \Fref{fig:zeromat}, we present zero-shot novel view synthesis results on mobile phone images captured directly by ourselves. We also show results using material-edited images generated with \cite{cheng2025zest}. These results demonstrate iIBR can synthesize novel views not only with various scene contents but also with different materials.

\noindent\textbf{Digital refocusing.}\quad
We introduce an interesting application of our 4D iIBR to photographic effects. In \Fref{fig:refocus}, we show the digital refocusing results.
We render the refocusing image after making dense light fields. 
Compared to the relevant works, Deepfocus~\cite{xiao2018deepfocus}, Deeplens~\cite{wang2018deeplens} and BokehMe~\cite{peng2022bokehme}, our 4D iIBR produces the realistic refocusing effect because the defocus blur is made from an integration of light rays over the lens aperture.
The background content is also visible through the defocus blur in our result which enhances the realism of refocusing, as it is generated and cast as the 4D ray.

\begin{figure}[t]
    \centering
     \includegraphics[width=\columnwidth]{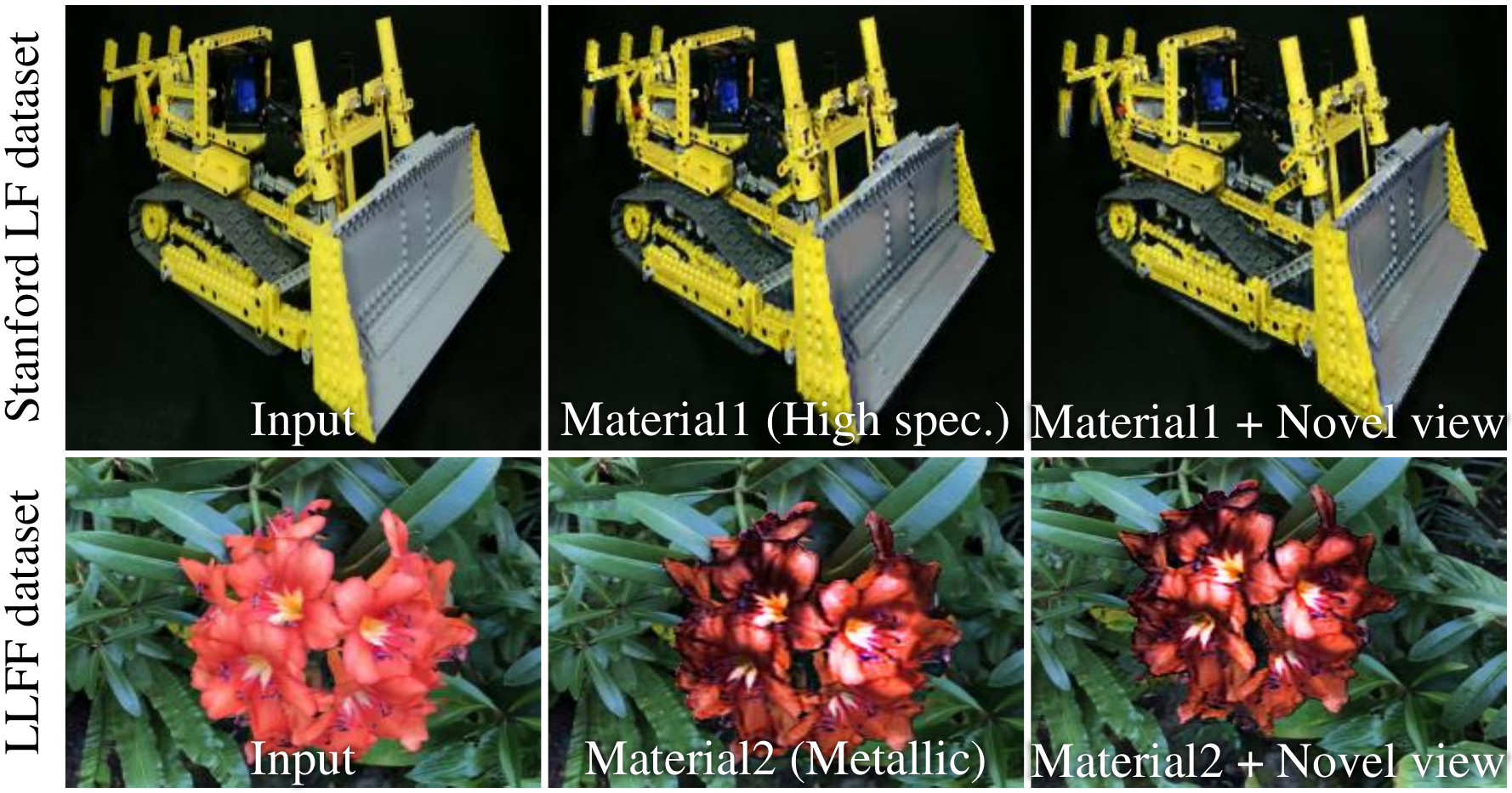}
     \vspace{-7mm}
     \caption{\textbf{Qualitative results for specularity simulation.} Our iIBR can synthesize novel views with specularity based on user-defined parameters.}
     \vspace{-5mm}
     \label{fig:general}
  \end{figure}

\noindent\textbf{Simulating specularity.}\quad
All novel views generated by iIBRnet assumes a Lambertian surface.
In \Fref{fig:general}, we report specularity simulation. Specular surfaces can be rendered through our iIBRnet if the BRDF and light source is provided (or defined by users).
The theoretical basis is as follows: First, specular surfaces appear as curved structures on the EPI as discussed in \cite{mildenhall2019local}, indicating that BRDFs can be synthesized on the EPI.
Second, we refer to the bottom-sided \Fref{fig:ray_embedding} again. By distorting the position of corresponding spatial rays on the EPI (the two gray-colored arrows), view-dependent effects can be rendered based on the rendering equation \cref{eq:render}, using the distorted weight from \cref{eq:4Dweight}, which is obtained by inputting ray embeddings from different positions than their original locations.


     \fontsize{10}{11.5}\selectfont

\vspace{-0.5mm}
\subsection{Quantitative results}
\vspace{-0.5mm}

\noindent\textbf{Novel view synthesis.}\quad
We evaluate the performance of our model on novel view synthesis. For this evaluation, we use the sub-aperture image from the center view as input and assess the quality of synthesized views over all other sub-aperture images. The result in \Tref{tab:quant} demonstrates that our method outperforms the comparision methods. We note that the performance drop for all methods on the NeRF LLFF dataset is due to the large baselines between target images. This highlights that unstructured viewpoints and large baselines pose significant challenges to the comparison works.

\noindent\textbf{Zero-shot cross validation.}\quad
To demonstrate that our 4D iIBR generalizes well across different datasets, we report zero-shot cross-validation results in \Tref{tab:zeroshot}. Our 4D iIBR consistently performs well regardless of the training data, thanks to our pixel-wise processing nature. However, there is a slight performance drop on the Stanford dataset due to specularities. The specularities make the ray transformer hard to synthesize the corresponding ray coordinates, even though the disparity positional encoding provides ray directional information for light flows. 


\noindent\textbf{Ablation study.}\quad
To evaluate the impact of our contributions, we conduct a series of ablation experiments on the Pov-ray dataset in \Tref{tab:ablation}. Key components of our iIBR, such as disparity positional encoding and inpainting, are tested, along with the loss terms, including the entropy loss and the EPI structure tensor loss. The results confirm that each component contributes to achieving better performance.
We also test several off-the-shelf depth foundation models in our pipeline. The result implies that our pipeline is effective with any model because all the models provide certain quality of output depth maps nowadays. Note that UniDepth~\cite{piccinelli2024unidepth} which offers approximate metric depth maps, is beneficial with repect to minimizing the need of learning scale/shift parameters.

   \fontsize{10}{11.5}\selectfont
  
\section{Conclusion}

We introduce a novel method to generate 4D light fields from single images, called inverse image-based rendering (iIBR). Through iIBR, we demonstrate that the inverse rendering of any light flow can be inferred from EPI pixel slope orientations. This insight allows us to generate and render continuous light flows of novel view images from single images. Additionally, we propose effective occlusion handling, enabling us to generate realistic rays in unseen areas. Through extensive evaluations, we show that our method outperforms recent novel view synthesis methods from single images and provides better generalization performance.

\noindent\textbf{Limitation and future direction.}\quad
Several directions exist for improving iIBR. One key challenge arises when moving objects are visible in scenes because of its temporal inconsistency of correspondences. Recent methods~\cite{lu20243d, wu20244d, park2021nerfies}, incorporating deformation fields, can be a good solution to reconstruct light flows of moving or deforming subjects over time.
Another limitation occurs when we attempt to render 360 degree images. This stems from the two-plane parameterization of light field photography, which causes the number of pixel colors that iIBRnet can reference to decrease as the significant viewpoint shifts away from the input image. Nevertheless, we believe that it is feasible by extending our approach to 360 geometry using a two-sphere parameterization, which is one of our future works.

\clearpage

\noindent\textbf{Acknowledgment} This work was supported by the National Research Foundation of Korea(NRF) grant funded by the Korea government(MSIT)(RS-2024-00338439) and Institute of Information $\&$ communications Technology Planning $\&$ Evaluation (IITP) grant funded by the Korea government(MSIT) (RS-RS-2021-II212068, Artificial Intelligence Innovation Hub and RS-2025-25441838, Development of a human foundation model for human-centric universal artificial intelligence and training of personnel)

{
    \small
    \bibliographystyle{ieeenat_fullname}
    \bibliography{main}
}

\end{document}